\documentclass{article}

\usepackage{microtype}
\usepackage{graphicx}
\usepackage{subfigure}
\usepackage{booktabs} 
\usepackage{subcaption}
\usepackage{amsthm}
\usepackage{amsmath}
\usepackage{amsfonts}
\usepackage[dvipsnames]{xcolor}
\theoremstyle{definition}
\newtheorem{definition}{Definition}[section]

\usepackage{hyperref}

\usepackage[accepted]{mlsys2024}

\mlsystitlerunning{Explainer-Agnostic Metrics}

\begin{document}

\twocolumn[
\mlsystitle{Evaluating Explainability in Machine Learning Predictions through Explainer-Agnostic Metrics}



\mlsyssetsymbol{equal}{*}

\begin{mlsysauthorlist}
\mlsysauthor{Cristian Munoz}{equal,hai}
\mlsysauthor{Kleyton da Costa}{equal,hai,puccs}
\mlsysauthor{Bernardo Modenesi}{equal,utah}
\mlsysauthor{Adriano Koshiyama}{hai}
\end{mlsysauthorlist}

\mlsysaffiliation{puccs}{Department of Computer Science, Pontifical Catholic University of Rio de Janeiro, Rio de Janeiro, Brazil}
\mlsysaffiliation{hai}{Holistic AI, London, UK}
\mlsysaffiliation{utah}{College of Health, University of Utah, Utah, USA}

\mlsyscorrespondingauthor{Cristian Munoz}{cristian.munoz@holisticai.com}

\mlsyskeywords{Machine Learning, MLSys}

\vskip 0.3in

\begin{abstract}
The rapid integration of artificial intelligence (AI) into various industries has introduced new challenges in governance and regulation, particularly regarding the understanding of complex AI systems. A critical demand from decision-makers is the ability to explain the results of machine learning models, which is essential for fostering trust and ensuring ethical AI practices. In this paper, we develop six distinct model-agnostic metrics designed to quantify the extent to which model predictions can be explained. These metrics measure different aspects of model explainability, ranging from local importance, global importance, and surrogate predictions, allowing for a comprehensive evaluation of how models generate their outputs. Furthermore, by computing our metrics, we can rank models in terms of explainability criteria such as importance concentration and consistency, prediction fluctuation, and surrogate fidelity and stability, offering a valuable tool for selecting models based not only on accuracy but also on transparency. We demonstrate the practical utility of these metrics on classification and regression tasks, and integrate these metrics into an existing Python package for public use.
\end{abstract}
]


\printAffiliationsAndNotice{\mlsysEqualContribution} 

\section{Introduction} 

Despite the remarkable recent evolution in prediction performance by artificial intelligence (AI) models, they are often deemed as ``black boxes'', i.e. models whose prediction mechanisms cannot be understood simply from their parameters. An explainable or interpretable algorithm is one for which the rules guiding its prediction decisions can be questioned and explained in a way that is intelligible to humans. Specifically, \textit{interpretability} regards the ability to extract causal knowledge about the world from a model, and \textit{explainability} pertains to the capability to articulate precisely how a complex model arrived at specific predictions, detailing its mechanics. Understanding AI models' behavior is essential for explaining predictions to support decision-making, debugging unexpected behaviors (contributing to improving model accuracy), refining modeling and data mining processes, verifying that model behavior is reasonable and fair, and effectively presenting predictions to stakeholders.

The main goal of explainable artificial intelligence (XAI) encompasses several critical objectives \cite{ali2023explainable}. XAI aims to empower individuals by enabling them to make informed decisions, mitigating the potential harms of fully autonomous decision-making systems. XAI also identifies and addresses vulnerabilities that could compromise machine learning-based systems, bolstering their resilience. Lastly, it boosts user confidence in AI systems by promoting transparency and fostering a more clear understanding of the decisions made by these models.

The AI literature offers various approaches to assessing explainability methods. Quantitative metrics can be used to assess whether these methods meet specific quality and reliability criteria \citep{bodria2023benchmarking}. Common metrics include fidelity—how well the explanation aligns with the underlying model \citep{guidotti2018survey}, stability—whether similar inputs yield consistent explanations \citep{alvarez2018towards, guidotti2019stability}, faithfulness—how accurately the explanation reflects the true behavior of the model \citep{alvarez2018towards}, monotonicity—whether more of a certain feature leads to a stronger explanation \citep{luss2021leveraging}, and complexity—how easily the explanation can be understood. Qualitative assessments are similarly varied and are categorized into functionally-founded, application-grounded, and human-grounded, each offering different perspectives on the utility and interpretability of the explainability method.

While the mentioned approaches primarily focus on evaluating the performance of explanation methods (i.e. explainers) rather than the actual explanations of the model’s behavior, this limits their generalizability and applicability in diverse real-world scenarios. For instance, \citet{rosenfeld2021better} proposes metrics such as the number of features and rules used in explanations to assess their simplicity and utility across models, emphasizing the need for stable and interpretable explanations. Furthermore, \citet{fel2022good} introduces generalizability and consistency measures to evaluate the robustness of explanations across different data subsets, ensuring that explanations do not conflict when models are applied to new, unseen data. Additionally, \citet{marek2023towards} focus on explanation stability under different types of noise and variations in training data, further contributing to the reliability of explanations, especially in models like linear classifiers and decision trees where stability plays a crucial role when data distributions shift. These approaches broaden the evaluative scope beyond just the explainers, offering a more comprehensive understanding of how explanations perform under real-world conditions.

This study introduces a novel set of explainer-agnostic metrics that can evaluate the outputs of any XAI method used for classification or regression tasks. These metrics encapsulate the behavior of explanations into a singular, concise representation, making them highly applicable in automated systems that monitor the trade-off between model accuracy and explainability. By quantifying this balance, the proposed metrics help assess both the transparency and risk associated with AI models in deployed environments.

Our main contributions are:

\begin{itemize}
    \item Explainer-Agnostic Metrics: We propose a set of explainer-agnostic metrics designed to evaluate the outputs of any explainer, making them applicable across various model types and settings.
    \item Consistent Feature Evaluation: Our metrics serve as a consistent framework for evaluating the feature importance of any explainer in deployment scenarios, ensuring uniformity for comparing different models and explainers.
    \item Publicly Available Tools\footnote{\url{https://github.com/holistic-ai/holisticai-research/tree/main/explainer_agnostic_metrics}}: All metrics and benchmark datasets are made publicly available via a Python library, providing the community with tools to implement and further develop these methods.
\end{itemize}

The paper is organized as follows: Section 2 presents our proposed methods for explainable AI at local and global levels; Section 3 discusses a set of applications for these methods; and finally, Section 4 outlines our main findings and potential directions for future research.


\section{Background and Literature}

The literature in explainable methods for AI models has been gaining prominence as AI models proliferate in virtually all areas of society -- such as predicting hypertension \citep{elshawi2019hypertension}, healthcare \cite{elshawi2020healthcare}, COVID-19 diagnosis \cite{buckmann2022interpretable}, economics and finance \cite{thimoteo2022explainable}. The discussion In this section is far from being exhaustive (for more details see Appendix A), given how effervescent this field is, but the goal is to motivate the reader about the XAI metrics literature and how our proposed metrics are positioned in this literature.

The absence of ground truth for explainability tools adds a layer of complexity to the comparison of different strategies, as pointed out by \citet{zhou2021metrics}. In order to circumvent this issue, authors assess XAI methods taking into account aspects such as fidelity, unambiguity, and overlap, as discussed by \citet{lakkaraju2017interpretable}. In fact, the existing literature introduces various methods for evaluating explainability techniques. \citet{mothilal2021unifying} presents a framework that unifies strategies centered around feature attribution and counterfactual generation. In contrast, other studies propose explainability metrics grounded in algorithmic stability \cite{fel2022good, khaire2022stability, nogueira2018stability}. However, it is essential to clarify that our objective is not to evaluate the explainability methods themselves. Instead, our focus is on providing insights based on the importance attributed to features by different models. In doing so, we aim to facilitate comparisons of explanations across models.

We lay out three categories of metrics with emphasis on the explainability results generated for AI models: (i) \textit{subjective}, (ii) \textit{objective}, and (iii) \textit{computational}. The \textit{subjective metrics} are employed when evaluating aspects that elicit subjective responses from users. This type of metric may be based on trust, understanding, and satisfaction, as proposed in \citet{hoffman2018metrics}. The \textit{objective metrics} are those related to observed aspects, for example, in users performing a particular task. These can be measures of time for the execution of the task or accuracy. \citet{schmidt2019quantifying} seeks to objectively measure the quality of explainability methods and shows that quick and highly accurate decisions represent a good understanding of explainability. \citet{narayanan2018human} evaluate explainability results based on subjective (satisfaction) and objective (response time and accuracy) metrics. Furthermore, \textit{computational metrics} are derived from mathematical indicators that assess the quality of explanations generated by an XAI method. Since these metrics are based on specific equations, user intervention is not necessary for obtaining them, making this type of metric suitable for automated systems.


\section{Proposed explainer-agnostic metrics}

The proposed metrics are model-agnostic and explainer-agnostic, meaning they can be applied to any type of model without requiring access to the model's internal structure or parameter estimates. These metrics only require the predictions $\hat{y}$ from a trained model $f$ and the explanations of any explainer $\mathcal{E}$ applied in $f$. Additionally, our methods utilize existing concepts in explainable AI (XAI), including (i) permutation feature importance, (ii) partial dependence plots, and (iii) surrogate models. By summarizing explainer outputs into a single value, proposed metrics provide a concise measure that captures essential behaviors of the explanations, enabling a consistent evaluation of the model's explanations across different settings.

\begin{figure*}[t]
    \centering
    \includegraphics[scale=0.35]{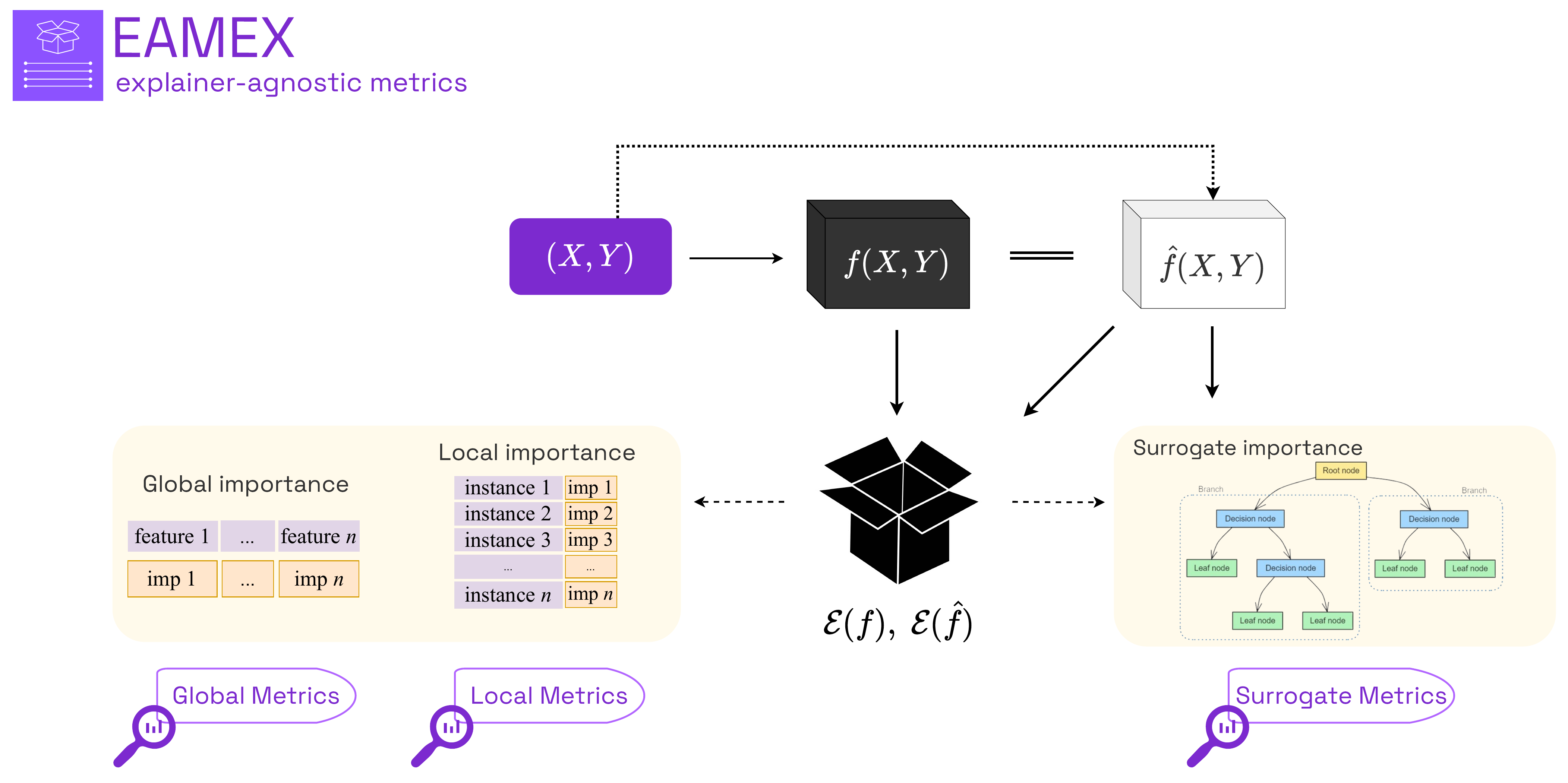}
    \caption{A simplified representation of explainer-agnostic metrics (EAMEX) framework}
    \label{fig:diagram}
\end{figure*}

Figure \ref{fig:diagram} illustrates a simplified framework for explainer-agnostic metrics. Given a set of features $X$ and a target vector $Y$, both the black-box model $f$ and its surrogate $\hat{f}$ serve as inputs to the explainer $\mathcal{E}$. The nature of $\mathcal{E}$ can vary, but we assume three possible types of outputs: (1) local feature importance focuses on understanding the contribution of features in the predictions of each observation; (2) global importance provides insights into the overall model prediction complexity, summarizing feature importance metrics across the entire dataset. The (3) surrogate importance involves approximating the complex model with a simpler, interpretable model (the surrogate model $\hat{f}$).

Following, we describe the explainer-agnostic metrics proposed.

\subsection{Metrics Based on Global Feature Importance}
\newcommand{\fis}{\textrm{S}}
Feature importance metrics are widely used by the XAI community, indicating the contribution of each feature to the model predictions. For this paper, we apply a simple normalization to the vector of feature importance values so it becomes a probability measure-- i.e. non-negative values that integrate up to one. We will refer to feature importance as this normalized vector henceforth throughout the paper.

\subsubsection{Feature Importance Spread}\label{sec:fis}

This metric evaluates the feature importance distribution using a divergence measure. Consider a uniform distribution of feature importance $U = \{\bar f, \bar f, \ldots, \bar f\}$, where $\bar f = 1/|F|$, indicating that all features are equally relevant. This distribution often suggests that understanding the model's decision-making process involves equally considering all (of the many) model features, making it often intractable for simple explanations to arise. 

We use this distribution $U$ as a benchmark for model complexity, quantifying deviations from it by using the Jensen-Shannon Divergence (JS), denoted as $\fis_{D}$. Unlike the Kullback-Leibler Divergence ($D_{KL}$), JS is symmetric and bounded between 0 and 1, making it a more interpretable measure of dissimilarity. A $\fis_{D}$ value close to 0 indicates that the feature importance distribution is near-uniform, implying that all features contribute similarly, making it harder to discern the key driving factors in the model. On the other hand, a higher $\fis_{D}$ value (closer to 1) indicates that feature importance is concentrated on a few features, which simplifies model interpretation by highlighting the most influential features.

\begin{definition}[Feature Importance Divergence]
Let $P$ be a normalized feature importance distribution, $U$ a uniform distribution, and $M = \frac{1}{2}(P + U)$. The feature importance spread is defined by

\begin{equation}
S_{D}(F) =\sum_{j = 1}^F \frac{1}{2} D_{KL}(P_{j} \| M_{j}) + \frac{1}{2} D_{KL}(U \| M_{j})
\end{equation}
\end{definition}

\subsubsection{$\alpha$-feature importance}

The $\alpha$-Feature Importance metric measures the smallest subset of features needed to represent at least $\alpha \times 100\%$ of the model's total feature importance.

\begin{definition}[$\alpha$-Feature Importance] Let $f_j$ represent the importance of the $j$-th feature, and $F$ be the total number of features. The $\alpha$-Feature Importance metric, $FI_{\alpha}(F)$, is the proportion of features required to capture at least $\alpha\times 100\%$ of the total importance:

\begin{equation}
    FI_{\alpha}(F) = \frac{ \min \left\{ k \, | \, \sum_{j=1}^{k} f_{(j)} \geq \alpha \cdot \sum_{j=1}^{F} f_{(j)} \right\} }{F}
\end{equation}

where $f_{(j)}$ are the ordered feature importances (from highest to lowest). $\alpha$ is the fraction of total importance you want to capture (e.g., $\alpha = 0.8$ for 80\%). $FI_{\alpha}(F)$ ranges from 0 to 1, indicating the proportion of features needed. A low $FI_{\alpha}(F)$ means a small number of features explain most of the model's behavior. Conversely, a high $FI_{\alpha}(F)$ means that many features are necessary to explain the model.

\end{definition}

\subsubsection{Fluctuation Ratio}

The Fluctuation Ratio ($F_R$) measures the oscillatory behavior in Partial Dependence Plots (PDPs). Each PDP denotes the model's average prediction for all the values of a chosen feature. Fluctuations in PDPs may indicate unstable or complex relationships between the feature and the target, complicating the task of explaining predictions.

\begin{definition}[Fluctuation Ratio]Given a model $f$ and feature $x$, the partial dependence function is $PD(x) = \hat{\mathbb{E}}_{\mathbf{X}_{-x}}[f(x, \mathbf{X}_{-x})]$, where $\mathbf{X}_{-x}$ represents the set of all features other than $x$. In order to compute the fluctuation ratio, the PDP is first interpolated over a grid for smoother derivative calculations. For each point $i$, the discrete derivative is $\Delta PD_i = PD(x_{i+1}) - PD(x_i)$, and $D_i = \operatorname{sign}(\Delta PD_i)$. The fluctuation ratio is:

\[
F_R = \frac{1}{n-2} \sum_{i=1}^{n-2} \mathbb{I}(D_i \neq D_{i+1}),
\]

where $\mathbb{I}$ is an indicator function that returns 1 when there is a change in direction ($D_i \neq D_{i+1}$), and 0 otherwise. Higher $F_R$ values indicate more frequent changes, suggesting more oscillations in the PDP and potentially less stability. For an overall fluctuation score across features, the average fluctuation ratio across $k$ features is:

\[
F_{R, \text{average}} = \frac{1}{k} \sum_{j=1}^{k} F_{R,j}.
\]

This average fluctuation ratio reflects the overall stability of the features' partial dependence plots without emphasizing the importance of individual features. It ranges from 0, when all PDPs are monotonically well behaved, to 1, when all of the PDPs are constantly changing the sign of their derivatives.
\end{definition}

\subsubsection{Rank Alignment}

The Rank Alignment metric quantifies how often the overall model feature importance ranking agrees with feature importance ranking for subsets (or subgroups) of the dataset. These subgroups can be e.g. different predicted classes in classification tasks or output quartiles in regression tasks.

\begin{definition}[Rank Alignment] 
Let $\mathbf{F}_\alpha$ be the top $\alpha$ proportion of features based on their importance in the overall dataset, $\mathbf{F}_\alpha^g$ be the set of top $\alpha$ proportion of features for group $g$, and $G$ be the total number of chosen groups. The Rank Alignment score is the average Jaccard similarities across all groups:

\begin{equation}
R_{A} = \frac{1}{G} \sum_{g=1}^{G} \frac{|\text{F}_\alpha \cap \text{F}_\alpha^g|}{|\text{F}_\alpha \cup \text{F}_\alpha^g|}
\end{equation}
\end{definition}

$R_{A}$ ranges from 1, when subgroups rankings fully agree with the global ranking, to 0, when the model explainability is better done within each subgroup.

\subsection{Metrics Based on Local Feature Importance}
\subsubsection{Rank Consistency}

The Rank Consistency metric evaluates how consistent the feature importance rankings are across different observations in the dataset. It measures the degree to which each feature maintains a consistent rank of importance when assessed at the local (observation-specific) level.

\begin{definition}[Rank Consistency]
For each observation $i$, the vector of local feature importances $f_i \in \mathbb{R}^d$ is converted into a ranking $r_i \in \mathbb{N}^d$, where $r_{ij} = \text{rank}(f_{ij})$. In this case, $r_{ij}$ represents the rank of feature $j$ in sample $i$, with lower ranks corresponding to higher importance. Next, for each feature $j$, the rank consistency is calculated based on the ranks it holds throughout the entire dataset. For each feature $j$, the most frequent rank $r_{\text{mode}, j}$ among iterations is determined, i.e. the mode of its ranking. Then, the average deviatio, $\bar{D}_j$, of the feature's ranks from this most frequent rank and the maximum observed deviation, $D_{\max, j}$, are computed as follows:

\begin{equation}
\begin{array}{rl}
\bar D_{j} = & \frac{1}{M}\sum_{i=1}^{M} |r_{ij} - r_{\text{mode}, j}| \\\\
D_{\max, j} = & \max_i \{r_{ij} \} - \min_i \{r_{ij}\}
\end{array}
\end{equation}

The rank consistency of feature $j$ is then calculated as:
\begin{equation}
C_{j} = 1 - \frac{\bar D_{j}}{D_{\max, j}}
\end{equation}

The overall rank consistency across all features is then given by:
\begin{equation}
R_{C} = \frac{1}{d} \sum_{j=1}^{d} C_j
\end{equation}
\end{definition}
$R_{C}$ is bounded by 0 and 1, with the higher values representing stable ranking for all observations, i.e. a model that has a consistent pattern in terms of feature importance.

\subsubsection{Importance Stability}

The Importance Stability ($I_S$) is a metric of \textit{relative} variability (or lack of stability) of the feature importance measures, averaged over all features, across the entire dataset. 

When feature importances are normalized such that their sum equals 1 for each observation (e.g., in SHAP values or other similar contexts), the stability of a feature's importance can be assessed by comparing its variance \textit{relative} to the maximum theoretical variance of random variable bounded between 0 and 1, with the same mean.

\begin{definition}[Importance Stability]
For each feature $j$, its mean importance across the samples is calculated as $\mu_j = \frac{1}{M} \sum_{i=1}^{M} f_{ij}$, where $f_{ij}$ is the importance of feature $j$ in sample $i$, and $M$ is the number of samples. The sample variance of the feature's importance is given by $V_j$:

\begin{equation}
V_j = \frac{1}{M} \sum_{i=1}^{M} (f_{ij} - \mu_j)^2
\end{equation}

To normalize the variance, we use the variance of a Bernoulli random variable, with similar mean, $\mu_j \times (1 - \mu_j)$, which represents the maximum possible variance for a feature when the importances are bounded between 0 and 1:

\[
V_{\text{max}, j} = \mu_j \times (1 - \mu_j)
\]

The stability $S_j$ for feature $j$ is calculated as:

\[
S_j = 1 - \frac{V_j}{V_{\text{max}, j}}  
\]

Finally, the overall stability across all features is calculated as:

\begin{equation}
    I_{S} =  \frac{1}{d} \sum_{j=1}^{d} S_j
\end{equation}

\end{definition}
$I_{S}$ ranges from 0 to 1, with higher values representing completely stable feature importance scores throughout the entire dataset, meaning the most easily explainable model.

\subsection{Metrics Based on Surrogate Models}

Surrogate models are simpler and interpretable models that approximate the behavior of complex black-box models, enabling interpretability for tasks such as feature importance analysis. These models are especially valuable in Explainable AI (XAI) as they allow predictions to be decomposed into understandable components. In order to ensure reliable model interpretation, it is crucial to evaluate the potential degradation caused by using the surrogate, i.e. the fidelity of the surrogate model, and the stability of feature importance across different samples.

\subsubsection{Performance Degradation}

The Performance Degradation metric measures how much the surrogate model's performance deviates from the original model. This metric applies differently to classification and regression tasks, depending on whether accuracy or error is being measured.

\begin{definition}[Performance Degradation]
Let $P_b$ represent the performance of the original model and $P_s$ the performance of the surrogate model.

- For classification, where $P_b$ and $P_s$ represent the accuracies of the original and surrogate models respectively, the degradation is calculated as:

\begin{equation}
D_{\text{classification}} = \frac{2(P_b - P_s)}{P_b + P_s}
\end{equation}

- For regression, where $P_b$ and $P_s$ are the mean squared errors (MSE) of the original and surrogate models, the degradation is:

\begin{equation}
D_{\text{regression}} = \max \left(0, \frac{2(P_s - P_b)}{P_b + P_s}\right)
\end{equation}

\end{definition}

Higher degradation values indicate a greater drop in performance from the surrogate model compared to the original model.

\subsubsection{Surrogate Fidelity}

The Surrogate Fidelity metric evaluates how closely the surrogate model replicates the predictions of the original model. Given the interpretability of the 3-level decision tree surrogate, we use accuracy for classification and normalized relative error for regression.

\begin{definition}[Surrogate Fidelity]
Let $y_{\text{pred}}$ and $y_{\text{surrogate}}$ represent the predictions of the original and surrogate models, respectively.

- For classification, fidelity is measured by accuracy:

\begin{equation}
F_{\text{classification}} = \frac{1}{n} \sum_{i=1}^{n} \mathbb{I}(y_{\text{pred},i} = y_{\text{surrogate},i})
\end{equation}

where $\mathbb{I}$ is the indicator function returning 1 if the predictions match and 0 otherwise.

- For regression, fidelity is measured by the normalized relative error:

\begin{equation}
F_{\text{regression}} = 1 - \frac{1}{n} \sum_{i=1}^{n} \frac{|y_{\text{pred},i} - y_{\text{surrogate},i}|}{\max(|y_{\text{pred},i}|, |y_{\text{surrogate},i}|)}
\end{equation}

A score closer to 1 indicates higher fidelity between the models.

\end{definition}

\subsubsection{Surrogate Feature Stability}

The Surrogate Feature Stability metric evaluates how consistently the surrogate model uses features across different bootstrapped samples. This is particularly useful to assess whether the surrogate model's feature selection is stable and reliable. The metric is applicable to both classification and regression tasks.

\begin{definition}[Surrogate Feature Stability]
Let $X$ be the feature matrix and $y_{\text{pred}}$ the predictions from the original model. The surrogate model is fitted on multiple bootstrapped samples, and the stability of the features is calculated by measuring the overlap of selected features across these samples.

- For both classification and regression, the feature stability is computed as:

\begin{equation}
F_{S} = \frac{1}{k} \sum_{i=1}^{k} \frac{|F_{\text{orig}} \cap F_{\text{resampled},i}|}{|F_{\text{orig}} \cup F_{\text{resampled},i}|}
\end{equation}

where $F_{\text{orig}}$ represents the features selected by the surrogate on the original data, $F_{\text{resampled},i}$ the features selected in the $i$-th bootstrap sample, and $k$ the number of bootstraps. The metric averages the Jaccard similarity between the original and resampled feature sets.

A score closer to 1 indicates higher stability in the surrogate model's feature selection across bootstrapped samples.
\end{definition}

\section{Experiments}

This section provides a detailed account of the experimental setup and presents the results derived from the application of explainability metrics to various machine learning models across standard benchmark datasets. The analysis is structured around two case studies: the first investigates the Adult dataset, while the second focuses on the US-Crime dataset. These case studies are designed to illustrate the effectiveness of the proposed explainability metrics in both classification and regression tasks. A comprehensive description of the datasets, model configurations, and explainability techniques employed is provided in the Appendix. 

\subsection{Case Study 1: Adult Dataset}

For the classification task, we employed the Adult Dataset considering training four classification models: Random Forest (RF), XGBoost (XGB), Logistic Regression (LR), and a Multi-Layer Perceptron (MLP).

\subsubsection{Global Feature Importance}

The analysis of Global Feature Importance metrics \textbf{alpha score} reveals that the majority of the models concentrate feature importance on approximately 13\% of the total features, as measured by the alpha score. An exception is the MLP model, which distributes its importance across roughly 20\% of the features. Moreover, XGB exhibits the highest \textbf{spread divergence} 0.75, indicating a stronger importance concentration in a few key features compared to other models. 

Additionally, the \textbf{fluctuation ratio}, a metric that quantifies the complexity of the relationship between features and predictions, is more pronounced in models with higher non-linear behavior, such as XGB (0.206) and RF (0.17). MLP shown a low fluctuation around 0.029. Conversely, LR, being a linear model, has the lowest fluctuation ratio close to 0, signifying that its predictions are more straightforward and interpretable (see Appendix B). Figure \ref{fig:adult_feature_importance_and_fluctuation_ratio_1} visualizes the top feature importances (in blue) along with their fluctuation ratios (in green). This visualization allows us to assess not only the importance of a feature for decision-making but also the complexity of its relationship with the target prediction. Features \textit{capital-loss}, \textit{hours-per-week} and \textit{age} shown the highest fluctuation ratio for XGBClassifier and RandomForestClassifier, with more than 30\% of the points changing their direction in the PDP plot.

Another noteworthy observation is the \textbf{rank alignment} of feature importance across predicted labels. In this regard, Linear Regression (LR) shows the highest alignment of features that contribute to both label 0 and label 1 predictions, demonstrating consistent usage in around 48.4\% of the features (see Table \ref{tab:classification_metrics}).

\begin{figure*}[t]
    \centering
    \subfigure[Permutation Feature Importance (Global Explainer) and Fluctuation Ratio for the ML model trained on the Adult Dataset.]{
        \includegraphics[width=0.95\textwidth]{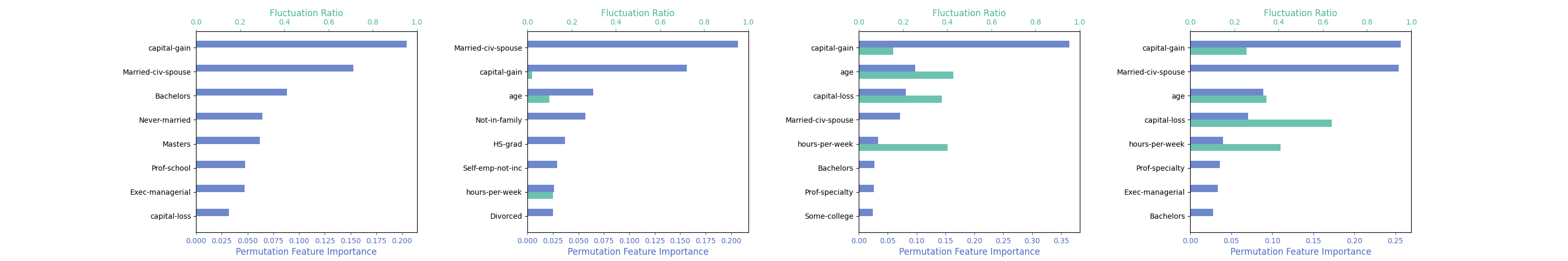}
        \label{fig:adult_feature_importance_and_fluctuation_ratio_1}
    }
    \vfill
    \subfigure[SHAP Importance (Local Explainer) and Importance Stability for ML models trained on Adult Dataset.]{
        \includegraphics[width=0.95\textwidth]{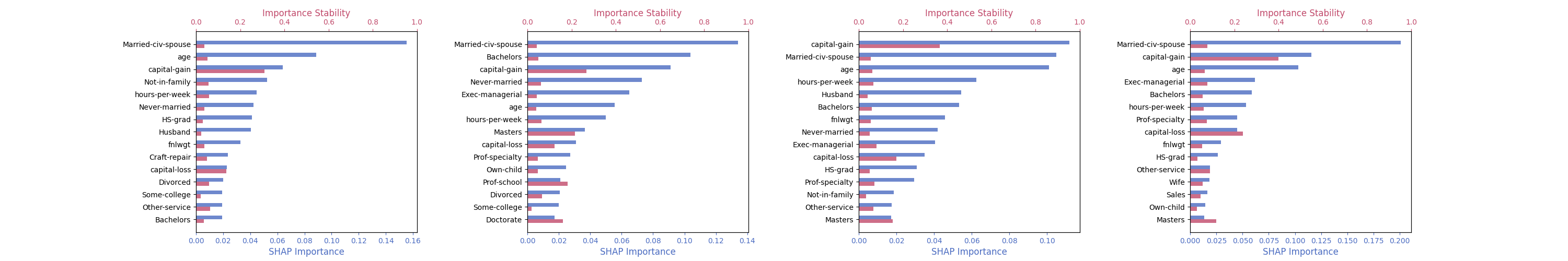}
        \label{fig:adult_feature_importance_and_fluctuation_ratio_2}
    }
    \vfill
    \subfigure[Feature Importance Contrast between samples grouped by labels for the ML model trained on the Adult Dataset. The upper half of the graph displays samples with label 0, and the lower half represents samples with label 1. The higher the \textcolor{blue}{blue} value, the further the feature importance of a specific instance is from the mode.]{
        \includegraphics[width=0.85\textwidth]{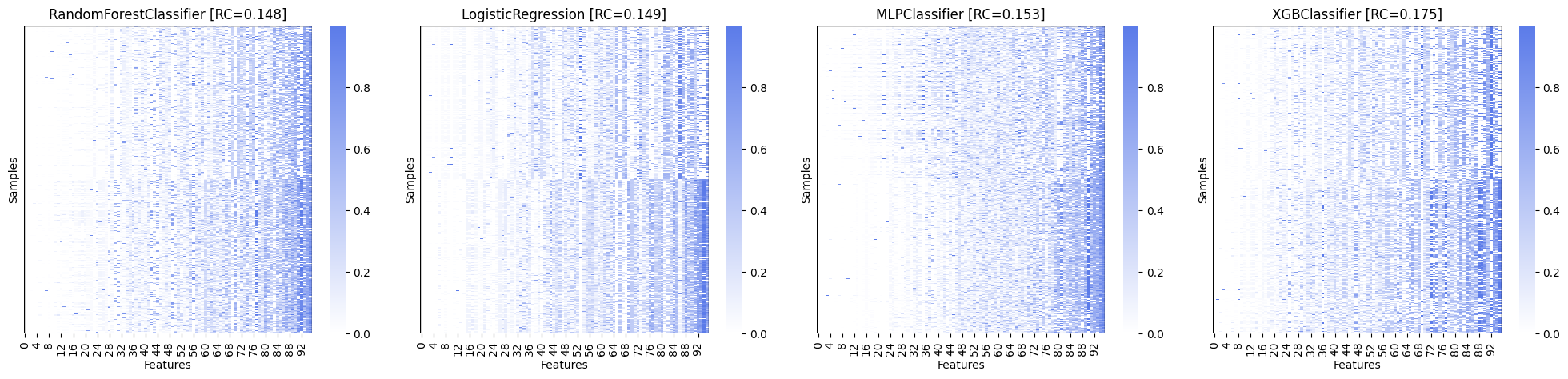}
        \label{fig:adult_position_consistency}
    }
    \caption{Comparison of different feature importance analyses on the Adult Dataset for an ML model.}
    \label{fig:adult_comparison}
\end{figure*}

\subsubsection{Local Feature Importance}

For \textbf{rank consistency}, derived from the SHAP explainer, Figure \ref{fig:adult_position_consistency} illustrates the consistency of feature rankings across all instances. The x-axis shows the features, while the y-axis represents the individual samples used to calculate local feature importances. The features are ordered by their consistency, with the most consistent features appearing on the right. The upper half of the figure corresponds to samples labeled as 0, and the lower half to samples labeled as 1.

This figure serves as a map of instability, where features on the left exhibit greater variability in ranking, primarily associated with samples labeled as 1 (lower half). Among the models tested, XGBoost displayed the highest variation, with a consistency score of 0.177, whereas Random Forest showed the lowest variation at 0.148.

For \textbf{importance stability}, this metric is a useful indicator of whether a feature's importance changes significantly across different instances. It helps to identify features that might be associated with more complex model decisions. In Figure \ref{fig:adult_feature_importance_and_fluctuation_ratio_2}, features like \textit{capital-gain}, \textit{capital-loss}, and \textit{Masters} exhibited higher instability across all models, with XGBoost showing the greatest instability, at around 40\%. This contributed to XGBoost achieving the highest instability score of 0.113.

\subsubsection{Surrogate Metrics}

The surrogate model analysis indicates that the Random Forest (RF) model exhibits the largest deviation from its surrogate, with a \textbf{surrogate fidelity} of 85.6\%, suggesting that the surrogate struggles to accurately replicate the original model’s predictions. Additionally, the use of surrogates results in \textbf{performance degradation}, with XGBoost and Random Forest showing the highest drops in accuracy, at 5.8\% and 2.6\%, respectively.

The \textbf{surrogate feature stability} metric provides insights into the consistency of feature importance when the surrogate model is generated by bootstrapping the dataset. For instance, the surrogate for the LR model demonstrated the highest feature stability at 31.71\%, while MLP had the lowest at 18.9\%. These percentages represent the proportion of features that persist across the bootstrapped surrogates.

\begin{table}
\footnotesize
\setlength{\tabcolsep}{2pt}
\begin{center}
\caption{Results XAI metrics for models trained on Adult Dataset}
\label{tab:classification_metrics}
\begin{tabular}{lccccc}
\toprule
\textbf{Metrics} & RF & XGB & LR & MLP & REF \\
\midrule
\textbf{Efficacy} &  &  &  &  &  \\
\midrule
Accuracy & 0.849 & 0.874 & 0.850 & 0.844 & 1 \\
F1-Score & 0.665 & 0.718 & 0.658 & 0.657 & 1 \\
\midrule
\textbf{Global Feature Imp.} &  &  &  &  &  \\
\midrule
Spread Divergence & 0.728 & 0.755 & 0.697 & 0.630 & 1 \\
Alpha Score & 0.125 & 0.083 & 0.125 & 0.208 & 0 \\
Fluctuation Ratio & 0.170 & 0.206 & 0.000 & 0.029 & 0 \\
Rank Alignment & 0.421 & 0.321 & 0.484 & 0.467 & 1 \\
\midrule
\textbf{Local Feature Imp.} & & & & & \\
\midrule
Rank Consistency & 0.148 & 0.175 & 0.149 & 0.153 & 0 \\
Importance Stability & 0.098 & 0.112 & 0.082 & 0.068 & 0 \\
\midrule
\textbf{Surrogate} & & & & &  \\
\midrule
Acc. Degradation & 0.041 & 0.065 & 0.039 & 0.032 & 0 \\
Surr. Fidelity & 0.856 & 0.876 & 0.895 & 0.865 & 1 \\
Surr. Feature Stability & 0.206 & 0.236 & 0.317 & 0.189 & 1 \\
\bottomrule
\end{tabular}
\end{center}
\end{table}

\subsection{Case Study 2: US-Crime Dataset}

For the regression task, we employed the US-Crime Dataset to train and evaluate four models: Random Forest (RF), XGBoost (XGB), Logistic Regression (LR), and Multi-Layer Perceptron (MLP). In terms of efficacy, the XGB and LR achieved the lowest root mean squared error (RMSE) at 0.138, closely followed by Random Forest with RMSEs of 0.141. The MLP model showed relatively poorer performance with the highest RMSE at 0.181.

\subsection{Global Feature Importance}

The global feature importance analysis reveals notable differences in how various models utilize features. Random Forest (RF) exhibits the highest \textbf{spread divergence}, at 0.723, significantly greater than the other models. Additionally, its \textbf{alpha score} indicates that 8.9\% of the features account for 80\% of the total importance. In contrast, the Multi-Layer Perceptron (MLP) has the lowest spread divergence, at 0.342, and requires 48.5\% of the features to account for 80\% of the total importance, suggesting a more balanced utilization of features across the dataset.

When examining the complexity of feature relationships, Random Forest and XGBoost demonstrate the highest \textbf{fluctuation ratios}, with values of 0.247 and 0.227, respectively, indicating more complex, non-linear relationships between features and predictions. For both Random Forest and XGBoost, the features \textit{state} and \textit{Pctlleg} show the highest fluctuations, with over 30\% fluctuation in the Partial Dependence Plot (PDP). In contrast, Logistic Regression shows no fluctuation, reflecting its linear nature, while the MLP has a minimal fluctuation ratio of 0.040.

Logistic Regression demonstrates the highest \textbf{rank alignment} for consistency across prediction ranges, with 70.6\% of features maintaining consistent importance rankings across the four output region subsets ($Q01$-$Q12$-$Q23$-$Q34$). This suggests that the importance of features remains stable regardless of the predicted value. In contrast, Random Forest exhibits the lowest rank alignment, with only 19.6\% of features retaining their importance across these subsets.

\begin{figure*}[t]
    \centering
    \subfigure[Permutation Feature Importance and Fluctuation Ratio for the ML model trained on the US-Crime Dataset.]{
        \includegraphics[width=0.95\textwidth]{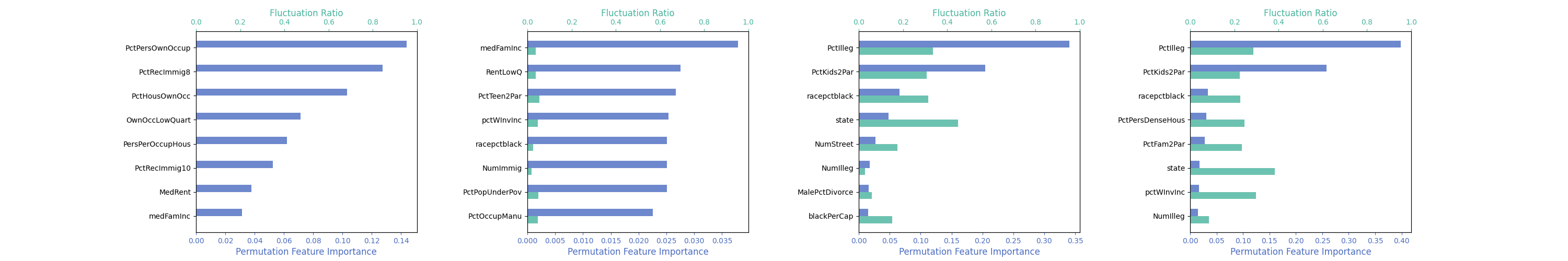}
        \label{fig:us_crime_feature_importance_and_fluctuation_ratio_1}
    }
    \vfill
    \subfigure[Local Feature Importance and stability for ML models trained on US-Crime Dataset.]{
        \includegraphics[width=0.95\textwidth]{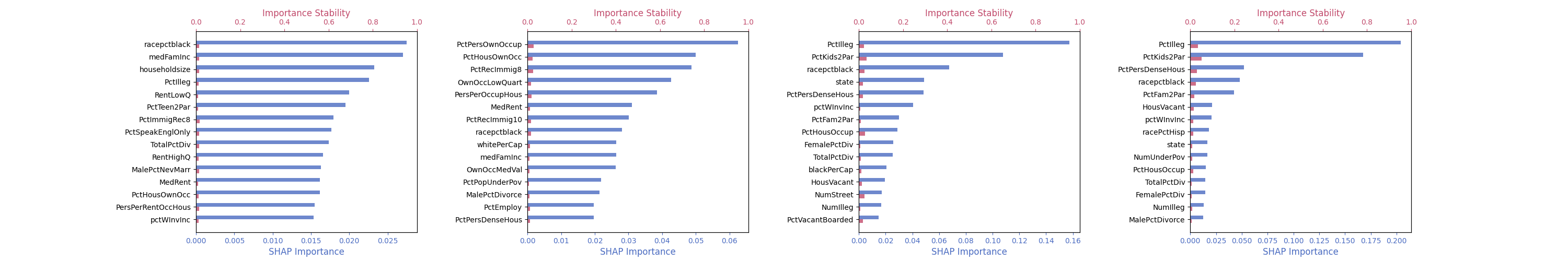}
        \label{fig:us_crime_feature_importance_and_fluctuation_ratio_2}
    }
    \subfigure[Feature Importance Contrast between samples grouped by labels for the ML model trained on the US-Crime Dataset. The higher the \textcolor{blue}{blue} value, the further the feature importance of a specific instance is from the mode.]{
        \includegraphics[width=0.85\textwidth]{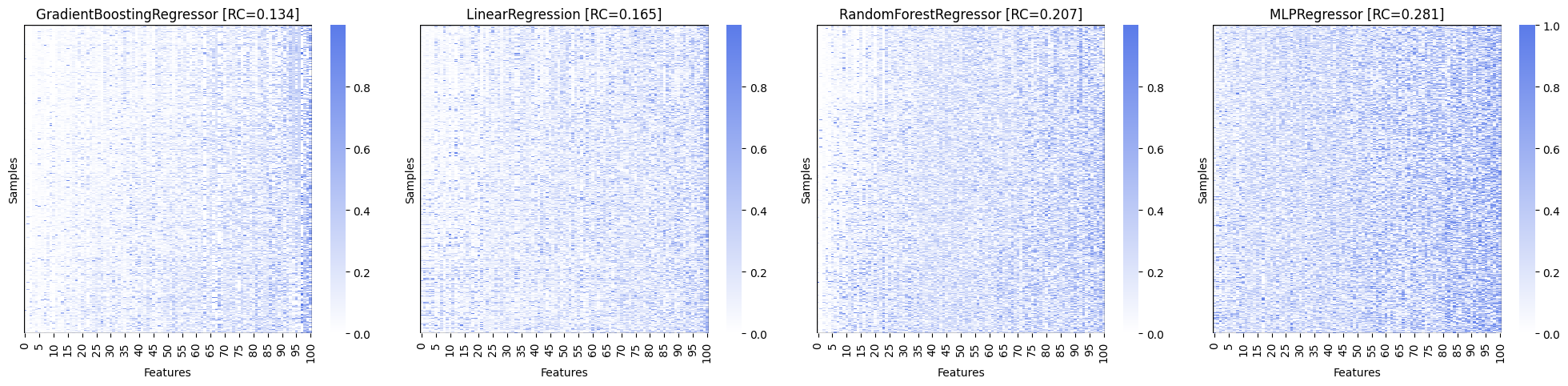}
        \label{fig:us_crime_position_consistency}
    }
    \caption{Comparison of different feature importance analyses on the US-Crime Dataset for an ML model.}
    \label{fig:us_crime_comparison}
\end{figure*}

\subsubsection{Local Feature Importance}

The local feature importance analysis offers valuable insights into how models behave at the individual sample level. For instances, XGBoost and Linear Regression demonstrate the best score for \textbf{rank consistencies} across the dataset, with values of 0.134 and 0.165, respectively. Figure \ref{fig:us_crime_position_consistency} illustrates the deviation of feature rankings from the mode for each sample in the dataset. From left to right, the plot reveals more frequent deviations in certain features (indicated by an increase in blue points), where greater deviations correspond to lower rank consistency.

\textbf{Feature stability} is generally low across all models, with a slight instability observed in the feature \textit{PctKids2Par} for both XGBoost and Random Forest, as illustrated in Figure \ref{fig:us_crime_feature_importance_and_fluctuation_ratio_2}.

\subsubsection{Surrogate Metrics}

The surrogate model analysis provides additional insights into model complexity and interpretability. XGBoost exhibits the highest \textbf{surrogate fidelity}, at 0.770, indicating that its behavior can be most effectively approximated by a decision tree (DT) model. In contrast, the Multi-Layer Perceptron (MLP) shows the lowest surrogate fidelity, at 0.480, suggesting more complex decision-making patterns.

In terms of \textbf{MSE degradation}, Random Forest and XGBoost perform the worst, with values of 0.272 and 0.228, respectively. MLP achieves the lowest degradation, with a value of 0.300, indicating less loss in predictive accuracy when approximated by the surrogate model. \textbf{Feature stability} is relatively consistent across models, with Linear Regression exhibiting the highest value of 67.4\%, reflecting the percentage of features that remain stable across different bootstrap samples.


\begin{table}
\footnotesize
\setlength{\tabcolsep}{2pt}
\begin{center}
\caption{Results XAI metrics for models trained on US-Crime Dataset}
\label{tab:regression_metrics}
\begin{tabular}{lccccc}
\toprule
\textbf{Metrics} & RF & XGB & LR & MLP & REF \\
\midrule
\textbf{Efficacy} &  &  &  &  &  \\
\midrule
RMSE & 0.141 & 0.138 & 0.138 & 0.181 & 0 \\
SMAPE & 0.239 & 0.234 & 0.286 & 0.403 & 0 \\
\midrule
\textbf{Global Feature Imp.} &  &  &  &  &  \\
\midrule
Spread Divergence & 0.723 & 0.700 & 0.671 & 0.342 & 1 \\
Alpha Score & 0.089 & 0.129 & 0.158 & 0.485 & 0 \\
Fluctuation Ratio & 0.247 & 0.227 & 0.000 & 0.040 & 0 \\
Rank Alignment & 0.196 & 0.279 & 0.706 & 0.426 & 1 \\
\midrule
\textbf{Local Feature Imp.} & & & & & \\
\midrule
Rank Consistency & 0.207 & 0.134 & 0.165 & 0.281 & 0 \\
Importance Stability & 0.023 & 0.018 & 0.011 & 0.011 & 0 \\
\midrule
\textbf{Surrogate} & & & & &  \\
\midrule
MSE Degradation & 0.272 & 0.228 & 0.300 & 0 & 0 \\
Surr. Fidelity & 0.763 & 0.770 & 0.674 & 0.480 & 1 \\
Surr. Feature Stability & 0.489 & 0.270 & 0.338 & 0.202 & 1 \\
\bottomrule
\end{tabular}
\end{center}
\end{table}

\subsection{Overall analysis for explainer-agnostic metrics}

\begin{figure*}[t]
    \centering
    \subfigure[Permutation Feature Importance and Fluctuation Ratio for the ML model trained on the Adult Dataset.]{
        \includegraphics[width=0.45\textwidth]{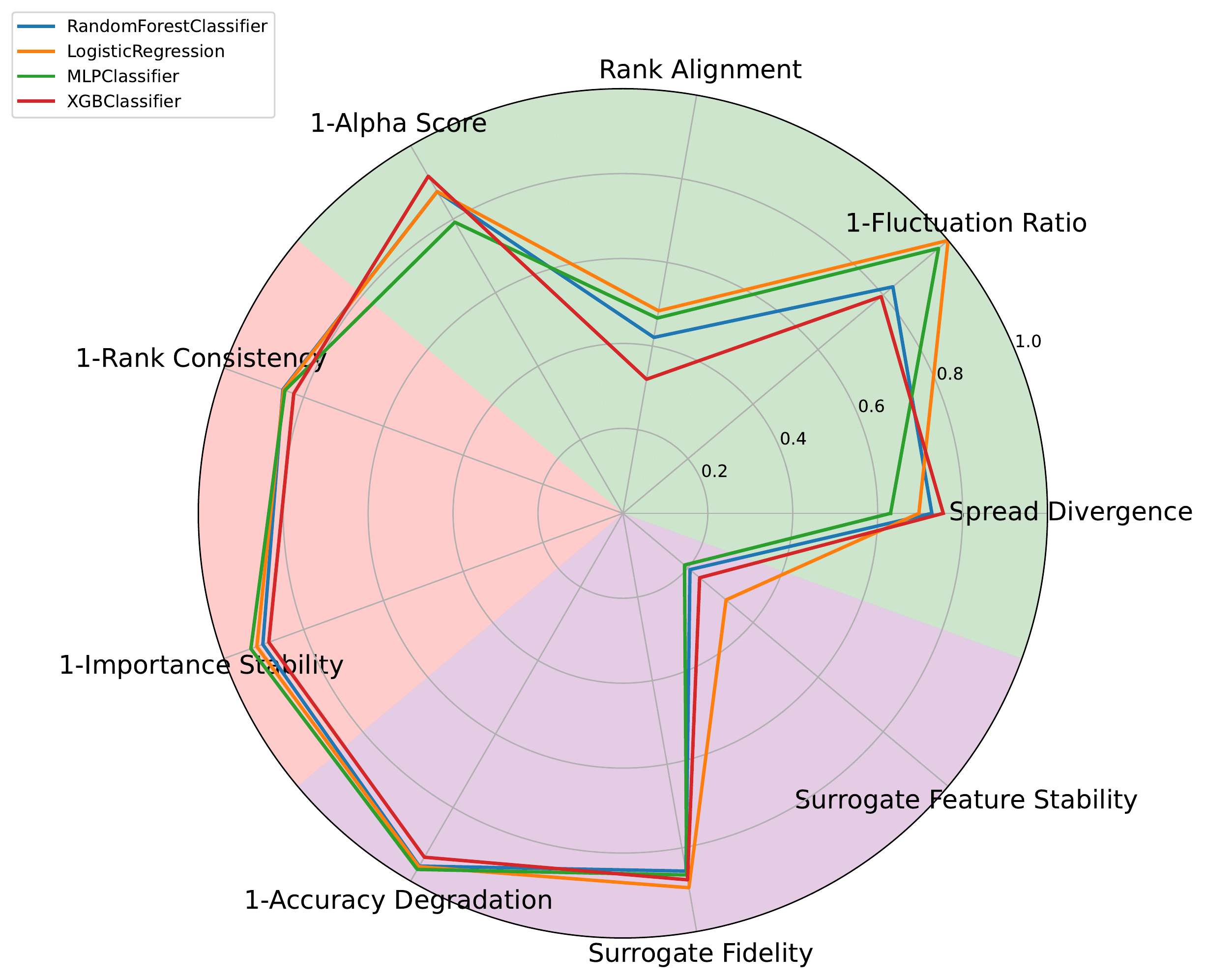}
        \label{fig:adult_radar}
    }
    \subfigure[Feature Importance Contrast between samples grouped by labels for the ML model trained on the Adult Dataset.]{
        \includegraphics[width=0.45\textwidth]{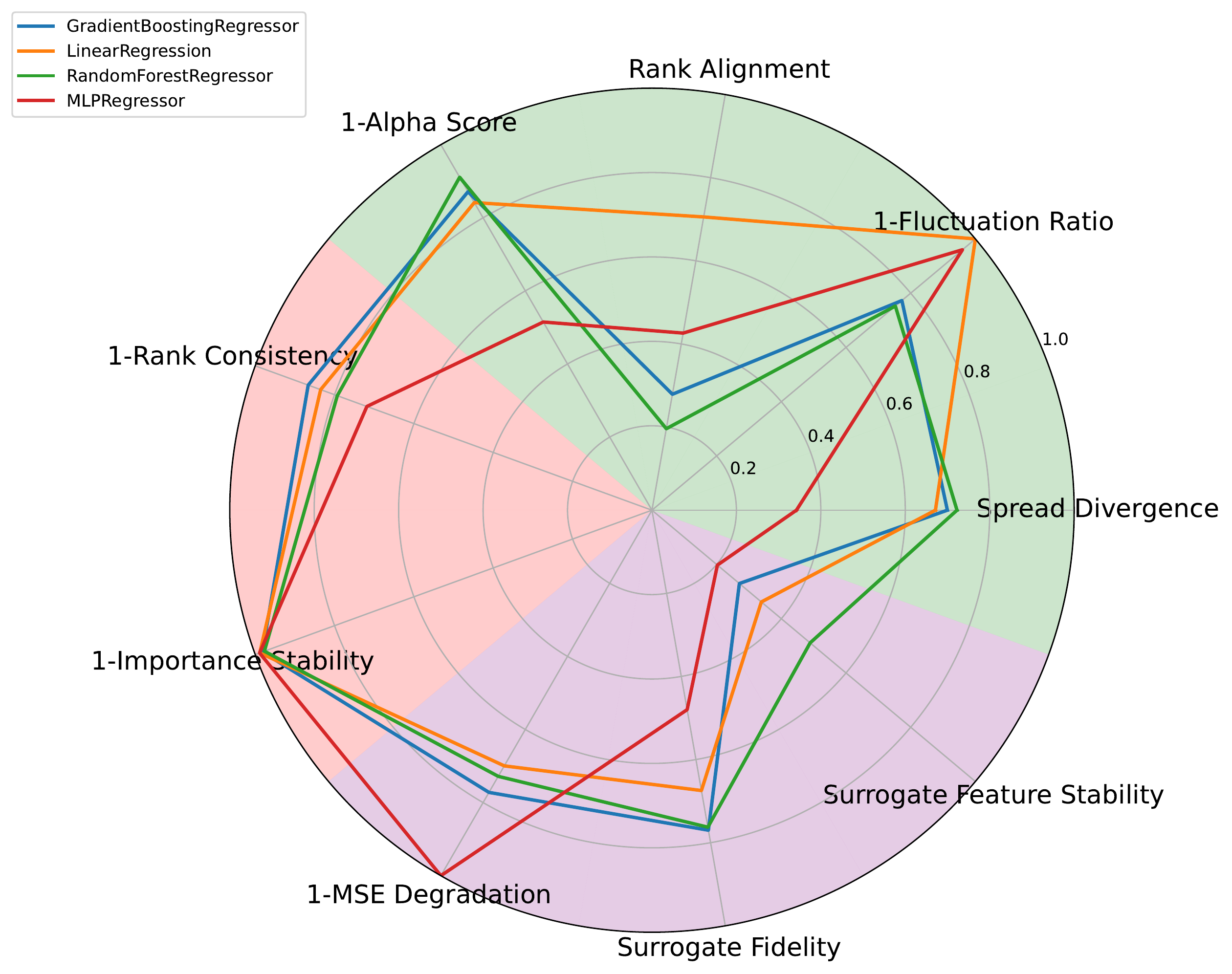}
        \label{fig:us_crime_radar}
    }
    \caption{Overall analysis of explainer-agnostic metrics for binary classification (a) and regression (b) tasks. The color areas represent \textcolor{OliveGreen}{global importance}, \textcolor{BrickRed}{local importance}, and \textcolor{Orchid}{surrogate importance} metrics. All metrics' reference values were standardized to facilitate interpretation, so a value of 1 is considered the reference or desired value for all metrics.}
    \label{fig:radar_comparison}
\end{figure*}

The overall analysis of the explainer-agnostic metrics is illustrated in Figure \ref{fig:radar_comparison}. This visual framework helps us interpret model performance from two complementary perspectives. The first aims to capture the geometry of the metric outcomes, which reflects the balance among different feature importance categories analyzed in this work—global, local, and surrogate. Ideally, models with uniform metric behavior will exhibit shapes closer to a regular polygon or a circle, indicating stable and consistent results across categories. Conversely, irregular geometries suggest imbalances across the metric groups, revealing potential trade-offs or weaknesses.

\paragraph{Binary Classification} In Case Study 1, we observe that models have a decrease in explainability considering rank alignment and surrogate feature stability. These two metrics emerges as a critical outlier, disrupting the overall geometry. As detailed in Table \ref{tab:classification_metrics}, all models perform poorly on this metric, suggesting limitations in maintaining a consistent ranking of feature importances across multiple explainers.

\paragraph{Regression Analysis} Case Study 2 highlights more diverse and non-uniform geometries across the evaluated models, suggesting greater variability in metric behavior for regression tasks. Linear Regression achieves robust performance across most metrics, although it suffers from a drop in surrogate fidelity, likely due to the linear model’s inherent simplicity. In contrast, more complex models such as Random Forest and XGBoost show comparable behavior, characterized by "heart-shaped" geometries. MLP shape reflects asymmetric performance, with good results in fluctuation ratio, importance stability, and rank consistency but poor performance in other metrics.

\paragraph{Model Monitoring and Practical Use Cases} A key takeaway from this analysis is the utility of explainer-agnostic metrics as a monitoring tool. These metrics are not only valuable for comparing different models and architectures, as demonstrated in our case studies but also for tracking the behavior of individual models across different versions during deployment. Monitoring performance drift, such as shifts in surrogate fidelity or rank alignment, can alert practitioners to potential concept drift or fairness issues in production. Furthermore, explainer-agnostic metrics offer a straightforward and interpretable way to evaluate model stability, supporting tasks like drift detection and ensuring that interpretability remains intact as models evolve. This capability is especially important in high-stakes environments (e.g. finance or healthcare) where stable and explainable behavior is paramount.

\section{Conclusion}

The increasing focus on transparency in AI systems has highlighted the trade-off between accuracy and explainability in machine learning models. This paper introduces a new set of explainability metrics aimed at summarizing the vast amount of information contained in global and local feature importance, PDPs and surrogate models. These metrics offer a structured way to understand model complexity without the need for extensive graphical analysis of feature importance, providing a more systematic and interpretable framework for evaluating AI models. It is crucial to note that these metrics are explainer-agnostic, being applied for any explainability method.

As for future research, we plan on exploring the statistical properties of the metrics proposed, potentially creating inference tools for model selection and model auditing based on these tools. While the scope of the article focused on constructing model-agnostic metrics to evaluate model predictions, future studies may explore the development of metrics for methods focused on inner interpretability.


\bibliography{references}
\bibliographystyle{mlsys2024}

%



\end{document}